\documentclass[conference]{IEEEtran}
\usepackage{cite}
\usepackage{amsmath,amssymb,amsfonts,amsthm}
\usepackage{algorithm}
\usepackage{algorithmic}
\usepackage{graphicx}
\usepackage{booktabs}
\usepackage{multirow}
\usepackage{url}
\usepackage{microtype}
\newtheorem{proposition}{Proposition}

\newcommand{\softmax}{\operatorname{softmax}}

\newcommand{\KL}{\operatorname{KL}}

\newcommand{\sg}{\operatorname{sg}}
\title{Efficient Linear Attention for Multivariate Time Series Modeling via Entropy Calibration}
\author{%
\IEEEauthorblockN{Mingtao Zhang\IEEEauthorrefmark{1}, Guoli Yang\IEEEauthorrefmark{1}, Zhanxing Zhu\IEEEauthorrefmark{2}, Mengzhu Wang\IEEEauthorrefmark{1}, Xiaoying Bai\IEEEauthorrefmark{1}}%
\IEEEauthorblockA{\IEEEauthorrefmark{1}Advanced Institute of Big Data, Beijing\\
\IEEEauthorrefmark{2}University of Southampton}%
}

\begin{document}

\maketitle

\begin{abstract}

Attention mechanisms are effective for multivariate time series modeling because they adaptively aggregate information across timestamps and variables. Their quadratic cost, however, limits their use in multivariate and spatio-temporal systems. Linear attention reduces this cost by replacing nonlinear attention mappings with forms that support associative computation. Replacing exponential weighting may weaken the amplification of high-score positions and further bias weakly selective distributions toward uniform aggregation.
We propose an entropy-calibrated linear attention module and instantiate it as the \emph{ELinFormer} forecaster. Our operator decomposes aggregation into a uniform component and a query-dependent correction, whose magnitude is calibrated by a query-specific temperature. We derive a second-order KL proxy from centered query-key scores. The Taylor expansion reduces the proxy to the squared-score energy divided by the squared temperature. This score energy can be computed from shared products of the centered keys without forming the full pairwise score matrix, and the same products support the linear attention output. We design a smooth temperature rule to calibrate the KL proxy: it keeps low-energy residuals vanishing, approximately calibrates moderate-energy queries to a shared KL level, and allows the proxy to continue growing for high-energy queries.
Experiments on five PEMS benchmarks show competitive forecasting accuracy. Controlled comparisons with vanilla attention and Linformer demonstrate a favorable accuracy--memory trade-off, while analysis on PEMS08 shows that the proposed rational temperature rule and query-specific calibration achieve lower MAE than a linear rule or shared-temperature alternatives. These results support entropy calibration as an effective mechanism for preserving selective deviations in linear attention.

\end{abstract}


\section{Introduction}

Attention mechanisms have become a central component of modern sequence modeling~\cite{attention} and multivariate time series forecasting~\cite{informer,Crossformer}. In spatio-temporal settings, attention can model dynamic dependencies among variables and timestamps without relying only on a fixed graph. This flexibility is useful for traffic forecasting~\cite{STGCN,DCRNN}, where temporal and cross-variable correlations can vary with time.

The main drawback of standard attention is its quadratic complexity in the number of tokens or nodes. For an attention axis of length $M$, each head constructs an $M\times M$ score matrix for every sample. When the attention axis is large, this matrix can dominate GPU memory and runtime. Existing efficient attention methods reduce this cost through low-rank projections~\cite{linformer}, sparse patterns~\cite{sparsetransformer}, or task-oriented structured sparsity~\cite{nlpsparse}. Linear attention is especially attractive because associative computation can replace pairwise aggregation with shared matrix products. However, directly replacing exponential weighting with a linear form can weaken the contrast between high- and low-score positions. When the original attention is already broad, as in the diagnostic below, this change may further bias aggregation toward uniformity. An effective linear replacement should therefore preserve both efficient computation and useful query-dependent deviations from uniform attention.

\begin{figure*}[!t]
    \centering
    \includegraphics[width=\linewidth]{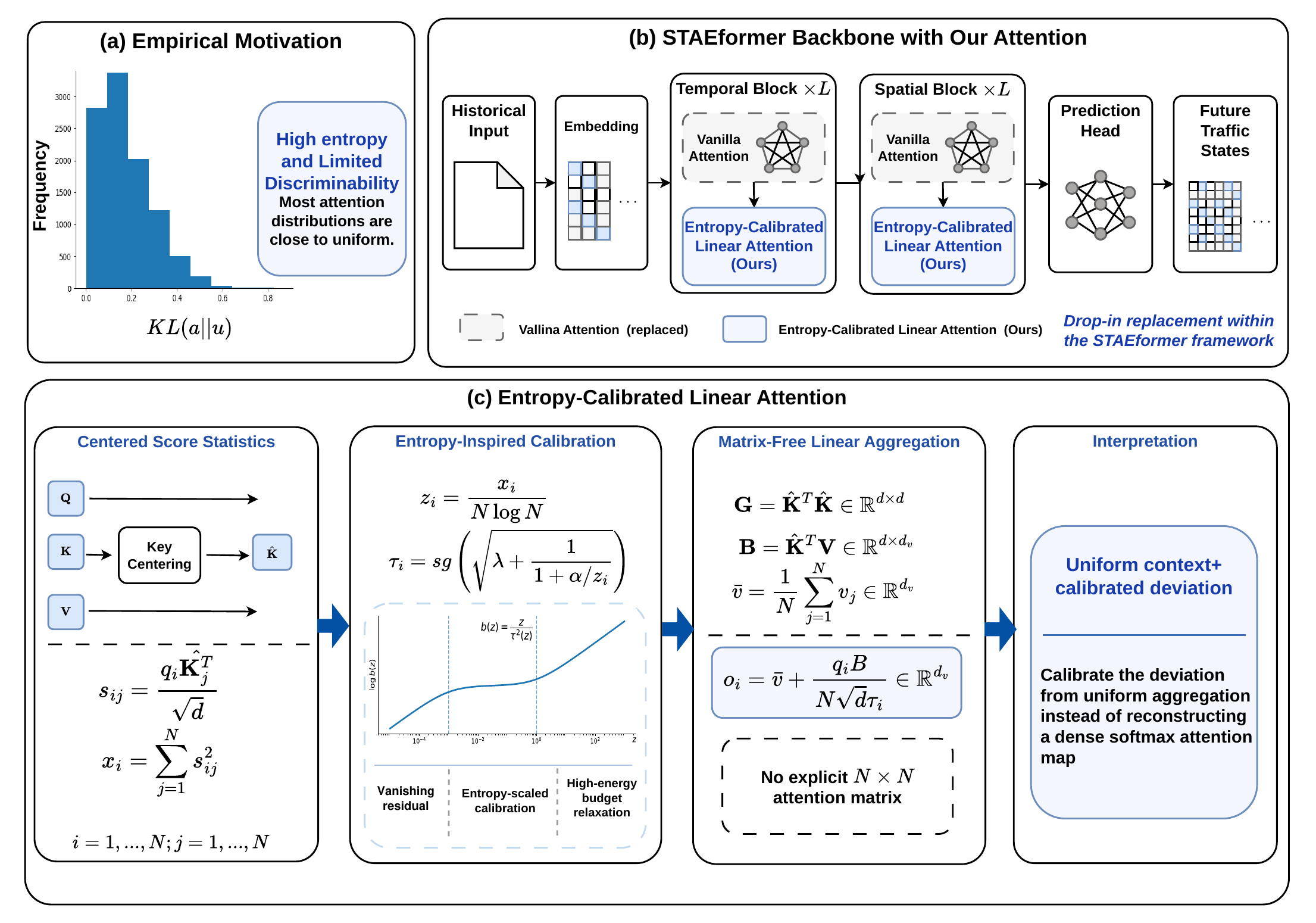}
    \caption{Motivation and overview. \textbf{(a) Illustrative diagnostic:} a representative PEMS08 histogram of $\KL(a_i\|u)$ from the vanilla STAEformer backbone is concentrated in the low-value region, where $a_i$ is a softmax attention vector over $M$ positions and $u=M^{-1}\mathbf{1}$ is the uniform attention. The result indicates high entropy and weak selectivity for many sampled queries. \textbf{(b) Backbone integration:} the embedding, feed-forward, normalization, and prediction components of STAEformer are retained, while its temporal and spatial attention modules are replaced by the proposed operator. \textbf{(c) Proposed module:} centered query-key scores yield the squared-score energy $x_i$. The normalized energy $z_i$ is mapped to a query-specific temperature $\tau_i$, and $b(z_i)=z_i/\tau_i^2$ is proportional to the KL proxy used for entropy calibration. Shared products of the centered keys then produce the large-axis linear attention output without forming the $M\times M$ pairwise attention matrix. The curve is smooth, and its labels describe the approximate behavior in low-, moderate-, and high-energy ranges.}
    \label{fig:framework}
\end{figure*}

Weakly selective or nearly uniform attention has been observed in several settings. Informer~\cite{informer}, for example, uses divergence from uniform to distinguish active queries from nearly uniform ones, while studies of attention entropy relate an excessive loss of entropy to training instability~\cite{entropy_drop}. Figure~\ref{fig:framework}(a) provides an additional task-specific example: many STAEformer~\cite{STAEformer} attention vectors sampled on PEMS08~\cite{pems} have small KL divergence from uniform attention. For a softmax attention vector $a_i$ over $M$ positions, this divergence equals $\log M-H(a_i)$, so a small value indicates high entropy and weak selectivity. While this does not imply that all attention vectors are uniform, it suggests that many queries use attention as a limited correction to a broad global context.

This perspective changes the approximation target. Instead of reconstructing every entry of a dense softmax matrix, we represent linear attention as uniform aggregation plus a query-dependent correction and control how far each query moves away from uniform attention. After centering the query-key scores, a second-order Taylor expansion gives a KL proxy determined by the squared-score energy and the squared temperature. The same proxy is proportional to the exact squared deviation of the linear coefficients from uniform mixing, linking entropy calibration to the strength of the query-dependent correction.

The squared-score energy can be evaluated from shared centered-key products in $O(Md^2)$ time, matching the complexity of the linear attention output. Since KL divergence from uniform attention is directly determined by attention entropy, controlling this proxy through the temperature also controls the estimated entropy. This relation motivates the term \emph{entropy calibration}.

A single fixed KL target can be too restrictive, especially for a linearized operator with weaker score contrast. For example, a clipped RMS rule assigns the same KL proxy to every query above a score-energy threshold and therefore removes differences among strong query--key patterns. We instead design a smooth temperature policy for three score-energy ranges. Low-energy residuals should vanish without unbounded gain, moderate-energy queries should remain near a shared KL level, and high-energy queries should not be forced to the same fixed level. A positive temperature floor and a bounded one-parameter transition satisfy these requirements without a hard threshold.

Based on this design, we propose an \emph{Entropy-Calibrated Linear Attention} (ECLA) module and refer to the resulting forecaster as \emph{ELinFormer}, as illustrated in Figure~\ref{fig:framework}. Centering the keys shifts all scores for a query by the same constant and leaves its softmax attention unchanged, while removing the first-order term in the expansion. More importantly, the resulting squared-score energy can be computed from a shared key product, which makes the KL proxy available without materializing the full pairwise score matrix. This score-energy view may also be useful for other efficient methods that need a low-cost estimate of attention selectivity. The final output combines a global context term with a calibrated query-dependent residual, and both terms can be computed from shared matrix products.

The contributions of this paper are threefold. First, motivated by the weak selectivity observed in prior work and illustrated in Figure~\ref{fig:framework}, we derive a second-order KL proxy that connects centered score energy, temperature, and estimated attention entropy. We further show how this proxy can be computed in $O(Md^2)$ time from shared products of centered keys, making the efficient computation reusable beyond the specific operator studied here. Second, we derive a smooth query-specific temperature rule from explicit requirements on low-, moderate-, and high-energy score patterns, and connect its calibration to exact residual value sensitivity and conditional row conditioning. Third, we demonstrate that the resulting forecaster, ELinFormer, achieves competitive accuracy on five public PEMS datasets and provides a favorable accuracy--efficiency trade-off relative to vanilla attention and Linformer.

\section{Related Work}\label{sec:related}
\textbf{Multivariate time series forecasting.}
Multivariate forecasting models aggregate information across temporal and variable dimensions. Representative approaches include recurrent graph models~\cite{DCRNN}, spatio-temporal graph convolutional models~\cite{STGCN}, and Transformer-based architectures~\cite{informer,Crossformer}. Our work focuses on making attention-based aggregation more efficient while retaining query-dependent interactions.

\textbf{Spatio-temporal graph forecasting.}
Spatio-temporal graph models use predefined graphs~\cite{STGCN,DCRNN} or learn graph structures~\cite{GTS,MTGNN,AGCRN,GWNet}. Dynamic dependency modeling is also explored by PDFormer~\cite{PDformer}. When pairwise dependency weights are computed from feature similarities, this modeling approach is closely related to self-attention. Our method follows this dynamic dependency view but replaces dense pairwise attention with a linear operator.

\textbf{Efficient attention.}
The quadratic cost of attention limits its scalability. Linformer~\cite{linformer} uses low-rank projections, while sparse Transformer variants~\cite{nlpsparse} evaluate only selected attention scores. Graph sparsification and spectral filtering can also reduce spatial computation~\cite{SGL,STWave}, and Informer~\cite{informer} reduces temporal attention cost by focusing on non-uniform queries. Associative linear attention replaces the pairwise matrix with feature-space products~\cite{katharopoulos2020}. These efficient methods introduce priors such as low rank, sparsity, or spectral locality. In contrast, our method retains centered dot-product scores and controls their effect through a query-specific temperature that can be computed without the full pairwise score matrix.

\textbf{Attention entropy and selectivity.}
Entropy measures whether an attention vector is broad or concentrated, while KL divergence from uniform attention measures its selectivity. Informer~\cite{informer} uses this criterion to distinguish active queries from nearly uniform
ones, and other work relates attention entropy to training stability~\cite{entropy_drop}. NaLaFormer restores query-norm-dependent spikiness through norm-direction decomposition and query-norm-aware feature maps~\cite{nalaformer}. PolaFormer models same- and opposite-signed interactions and uses a learnable power function to reduce the entropy of nonnegative linear attention~\cite{polaformer}. We estimate KL divergence from centered score energy and use it to determine the temperature of a signed linear attention operator. Its statistic $q_iGq_i^\top/d$ depends jointly on the current query and key geometry. The operator analysis connects this query-wise calibration with partial row equilibration, following the matrix-scaling perspective~\cite{diamond_boyd}.

\section{Problem Formulation}
\label{sec:problem}

We consider multivariate spatio-temporal forecasting. Let $X_{t}\in\mathbb{R}^{N\times F}$ denote observations at time $t$, where $N$ is the number of variables or sensors and $F$ is the number of features per variable. Given a historical window $\mathcal{X}_{t-L+1:t}=(X_{t-L+1},\ldots,X_t)$ of length $L$, the goal is to predict a future window $\mathcal{Y}_{t+1:t+H}=(X_{t+1},\ldots,X_{t+H})$ of horizon $H$. A forecasting model learns
\begin{equation}
    f_\Theta: \mathbb{R}^{L\times N\times F}\rightarrow \mathbb{R}^{H\times N\times F'},
\end{equation}
where $F'$ is the number of predicted target channels.

Attention-based forecasters embed the input into hidden states $Z\in\mathbb{R}^{L\times N\times C}$ and alternately aggregate information along the temporal and variable dimensions. Temporal attention operates over the $L$ historical timestamps for each node, while spatial attention operates over the $N$ nodes for each timestamp. Since $L$ is usually short and $N$ can be large, spatial attention is often the main computational bottleneck.

For one attention head and one attention axis of length $M$, let $Q,K,V\in\mathbb{R}^{M\times d}$ be the query, key, and value matrices. Here, $M=N$ for spatial attention and $M=L$ for temporal attention. Standard attention computes $\softmax(QK^\top/\sqrt d)V$. Our goal is to retain query-dependent score ranking and entropy control while avoiding the dense $M\times M$ attention matrix.

\section{Methodology}
\label{sec:method}

For one attention head, standard attention computes
\begin{equation}
    a_{ij}=\frac{\exp(q_i k_j^\top/\sqrt d)}{\sum_{\ell=1}^{M}\exp(q_i k_\ell^\top/\sqrt d)},
    \qquad
    o_i=\sum_{j=1}^{M}a_{ij}v_j.
\end{equation}
This operation costs $O(M^2d)$ time and $O(M^2)$ attention memory per head.

\subsection{KL Proxy Definition}
\label{sec:kl_proxy}

We first center the keys along the attention axis,
\begin{equation}
    \widehat k_j=k_j-\bar k,
    \qquad
    \bar k=\frac{1}{M}\sum_{\ell=1}^{M}k_\ell,
\end{equation}
and define scaled centered scores
\begin{equation}
    s_{ij}=\frac{q_i\widehat k_j^\top}{\sqrt d},
    \qquad
    \sum_{j=1}^{M}s_{ij}=0.
    \label{eq:centered_scores}
\end{equation}
Centering subtracts the same value $q_i\bar k^\top/\sqrt d$ from every score associated with query $i$. It therefore leaves the corresponding softmax attention vector unchanged.

To connect temperature with entropy, we apply softmax to the same centered scores with a query-specific temperature,
\begin{equation}
    p_{ij}(\tau_i)=
    \frac{\exp(s_{ij}/\tau_i)}{\sum_{\ell=1}^{M}\exp(s_{i\ell}/\tau_i)}.
    \label{eq:softmax_for_analysis}
\end{equation}
This softmax vector is used only for the following analysis and is not computed by the deployed linear module. Let $u=M^{-1}\mathbf 1$. Its KL divergence from uniform attention satisfies
\begin{equation}
    \KL(p_i(\tau_i)\|u)=\log M-H(p_i(\tau_i)).
    \label{eq:kl_entropy_identity}
\end{equation}
Using the zero-sum property in Eq.~\eqref{eq:centered_scores}, a second-order Taylor expansion around uniform attention gives
\begin{equation}
    \KL(p_i(\tau_i)\|u)
    =\frac{x_i}{2M\tau_i^2}
    +O\!\left(\left\|s_i/\tau_i\right\|^3\right),\quad x_i=\sum_{j=1}^{M}s_{ij}^2.
    \label{eq:kl_second_order}
\end{equation}
We call
\begin{equation}
    \widehat{\mathrm{KL}}_i=\frac{x_i}{2M\tau_i^2}
    \label{eq:kl_proxy}
\end{equation}
the \emph{KL proxy}. The corresponding entropy estimate is $\widehat H_i=\log M-\widehat{\mathrm{KL}}_i$. Thus, choosing $\tau_i$ to control the KL proxy directly controls the estimated attention entropy. The quantity $x_i$ is the exact squared-score energy, and we take it as an approximation to KL divergence only when applying the second-order expansion.

\subsection{Query-Specific Temperature}

Building on the KL proxy in Eq.~\eqref{eq:kl_proxy}, we normalize the score energy as
\begin{equation}
    z_i=\frac{x_i}{M\log M}.
    \label{eq:normalized_energy_final}
\end{equation}
The factor $\log M$ follows the scale of the KL proxy and makes the statistic easier to compare across attention lengths.

Motivated by the use of temperature to control attention entropy~\cite{entropy_drop}, we design the temperature policy from three requirements. First, a positive lower bound prevents unbounded amplification as score energy tends to zero. Second, it should change smoothly and remain bounded as the score energy grows. Third, a single parameter should control the transition from the low-energy range to stronger query-specific aggregation. Let $r(z)$ denote the bounded part of the temperature rule. We require $r(0)=0$, $r(z)\rightarrow1$ as $z\rightarrow\infty$, and $r(\alpha)=1/2$, where $\alpha>0$ sets the transition scale. The simplest first-order rational function satisfying these conditions is
\begin{equation}
    r(z)=\frac{1}{1+\alpha/z}=\frac{z}{z+\alpha},
\end{equation}
where the latter form is numerically more stable. The resulting temperature is
\begin{equation}
    \tau_i=\beta\sg\!\left(
    \sqrt{\lambda+\frac{z_i}{z_i+\alpha}}
    \right),\qquad \beta>0,
    \label{eq:temperature_final}
\end{equation}
where $\lambda>0$ is the temperature floor, $\alpha$ controls the transition scale, $\beta$ is a global temperature scale, and $\sg$ denotes stop-gradient. The scale adapts to the current query in the forward pass, while the gradient is not propagated through the query-dependent temperature computation.

At the level of the normalized attention scores, $\beta$ is scale-redundant with the query projection and $\alpha$. For any $c>0$, replacing $(q_i,\alpha,\beta)$ with $(cq_i,c^2\alpha,c\beta)$ leaves both $z_i/(z_i+\alpha)$ and the normalized score $s_{ij}/\tau_i$ unchanged. Equivalently, any setting with $\beta>0$ can be represented with $\beta=1$ by rescaling the query projection and $\alpha$. We therefore fix $\beta=1$ in all experiments, while retaining it in Eq.~\eqref{eq:temperature_final} to make the global scale explicit.

To describe how the temperature controls the KL proxy, define
\begin{equation}
    b(z)=\frac{z}{\tau^2(z)}=\frac{z}{\beta^2\left(\lambda+z/(z+\alpha)\right)}.
\end{equation}
For $z>0$, its derivative is
\begin{equation}
 b'(z)=
 \frac{(1+\lambda)z^2+2\lambda\alpha z+\lambda\alpha^2}
 {\beta^2\big((1+\lambda)z+\lambda\alpha\big)^2}>0.
\label{eq:b_derivative}
\end{equation}
Equation~\eqref{eq:kl_proxy} can then be written as $\widehat{\mathrm{KL}}_i=(\log M/2)b(z_i)$, which is a smooth function. When $\lambda$ is small, it has three useful score-energy ranges:
\begin{equation}
    b(z)\approx\frac{1}{\beta^2}
    \begin{cases}
        z/\lambda,
        & z\ll \dfrac{\lambda\alpha}{1+\lambda},\\[2mm]
        \alpha/(1+\lambda),
        & \dfrac{\lambda\alpha}{1+\lambda}\ll z\ll\alpha,\\[2mm]
        z/(1+\lambda),
        & z\gg\alpha.
    \end{cases}
    \label{eq:energy_ranges_final}
\end{equation}
In the low-energy range, the KL proxy approaches zero and the estimated entropy remains high. In the moderate-energy range, queries are approximately calibrated to a shared KL level. In the high-energy range, the temperature approaches a finite ceiling and the KL proxy can continue to grow, rather than being clipped to the same value for all strong queries. The temperature is bounded by
\begin{equation}
    \beta\sqrt\lambda\leq\tau_i<\beta\sqrt{1+\lambda}.
\end{equation}
Moreover, $b(z)$ is strictly increasing for $z>0$, so stronger squared-score energy always permits at least as much deviation from uniform attention.

\subsection{Linear Aggregation from Shared Products}

Using the calibrated temperature, we define the linear attention coefficients
\begin{equation}
    \widetilde a_{ij}=\frac{1}{M}\left(1+\frac{s_{ij}}{\tau_i}\right).
    \label{eq:linear_coefficients_final}
\end{equation}
Since the centered scores sum to zero, $\sum_j\widetilde a_{ij}=1$. The coefficients $\widetilde a_{ij}$ may be negative and are therefore signed mixing coefficients rather than probabilities. Their deviation from uniform attention has the exact form
\begin{equation}
    M\left\|\widetilde a_i-u\right\|_2^2
    =\frac{x_i}{M\tau_i^2}
    =2\widehat{\mathrm{KL}}_i.
    \label{eq:linear_kl_link}
\end{equation}
Thus, the same KL proxy that estimates softmax selectivity also controls the exact squared deviation of the linear coefficients from uniform mixing.

The output is
\begin{equation}
\begin{aligned}
    o_i
    &=\sum_{j=1}^{M}\widetilde a_{ij}v_j\\
    &=\bar v+\frac{1}{M\tau_i}\sum_{j=1}^{M}s_{ij}v_j,
    \qquad
    \bar v=\frac{1}{M}\sum_{j=1}^{M}v_j.
\end{aligned}
\label{eq:linear_output_final}
\end{equation}
The first term is the uniform value context, and the second is a query-dependent signed residual.

Both the score energy and the output can be computed from shared products of the centered keys. Let
\begin{equation}
    G=\widehat K^\top\widehat K,
    \qquad
    B=\widehat K^\top V.
\end{equation}
Then
\begin{equation}
    x_i=\frac{1}{d}q_iGq_i^\top,
    \qquad
    o_i=\bar v+\frac{q_iB}{M\sqrt d\,\tau_i}.
    \label{eq:shared_product_output}
\end{equation}
The same centered-key products therefore support both entropy calibration and linear aggregation without materializing the pairwise score matrix. Since the proxy computation does not depend on the specific linear output form, this observation may also be useful in other efficient attention methods that need a low-cost estimate of attention selectivity.

\subsection{Query-wise Calibration and Operator Sensitivity}
\label{sec:operator_sensitivity}
The bounded temperature in ECLA calibrates the signed residual scale while retaining variation among strong query--key patterns, addressing the weaker amplification of large positive scores by the affine response.
The entropy-derived statistic also measures the sensitivity of the deployed linear operator. Fix $Q,K$, $M>1$, and positive $\alpha,\beta,\lambda$. Write $S=Q\widehat K^\top/\sqrt d$, $D_\tau=\operatorname{diag}(1/\tau_i)$, and $\mathcal J_\tau=D_\tau S/M$. For one projected value channel $v\in\mathbb R^M$, the signed residual is $r=\mathcal J_\tau v$. Since $x_i=\lVert s_i\rVert_2^2$, its row-wise value sensitivity is exactly
\begin{equation}
\left\lVert\frac{\partial r_i}{\partial v}\right\rVert_2^2
=[\mathcal J_\tau\mathcal J_\tau^\top]_{ii}
=\frac{x_i}{M^2\tau_i^2}
=\frac{2\widehat{\mathrm{KL}}_i}{M}.
\label{eq:operator_sensitivity}
\end{equation}
This identity concerns projected value activations with fixed queries and keys and holds independently of the near-uniform Taylor approximation. The approximation of the true softmax KL by $\widehat{\mathrm{KL}}_i$ uses the local expansion in Section~\ref{sec:kl_proxy}; the value Jacobian here is distinct from the full $W_Q,W_K,W_V$ parameter Jacobian.

The identity also gives a conditional explanation for query-wise calibration. Let $I$ be any subset of at least two linearly independent rows of $S$, let $C_{ij}=\langle s_i,s_j\rangle/\sqrt{x_i x_j}$ for $i,j\in I$, and define $J_0=S_{I,:}/M$, $J_\tau=(D_\tau S)_{I,:}/M$, and $\Gamma_0=J_0J_0^\top$, $\Gamma_\tau=J_\tau J_\tau^\top$. These are output-space Gram matrices of the residual value Jacobian. If $z_{\min}$ and $z_{\max}$ are the normalized-energy extrema in $I$, then:
\begin{proposition}[Conditional value-space conditioning]
\label{prop:row_conditioning}
For the above linearly independent row subset,
\begin{equation}
\begin{aligned}
\frac{\kappa(\Gamma_\tau)}{\kappa(\Gamma_0)}&\leq \kappa(C)\,\chi_\tau,\\
\chi_\tau&=\frac{\lambda+z_{\min}/(z_{\min}+\alpha)}{\lambda+z_{\max}/(z_{\max}+\alpha)}.
\end{aligned}
\label{eq:conditioning_bound}
\end{equation}
Consequently, $\kappa(C)\chi_\tau<1$ is a sufficient condition for strict improvement on this local subsystem.
\end{proposition}
Here $\kappa$ is the spectral condition number. Supplement Section~S11 proves the bound using a diagonal factorization and Rayleigh quotients. The subset $I$ is an analysis device; the implementation processes every query, while $\operatorname{rank}(S)\leq\min(d,M-1)$ limits the full row Gram. Positive shared scaling only rescales the Gram and leaves its condition number unchanged. Query-wise scaling instead provides operator-level partial row equilibration under the stated sufficient condition.

Temperature scaling preserves each row's direction, so $C$ isolates directional correlations from the energy span. Calibration compresses that span while retaining the interaction geometry. The proposition states a sufficient balance between these two factors, providing a conditioning interpretation of query-wise calibration.

The same statistic preserves the linear implementation. With $G=\widehat K^\top\widehat K$, $B=\widehat K^\top V$, and $D_\tau=\operatorname{diag}(1/\tau_i)$, the output is
\begin{equation}
O=\mathbf 1_M\bar v+\frac{(D_\tau Q)B}{M\sqrt d}.
\label{eq:matrix_output}
\end{equation}
The score statistic supplies an entropy interpretation, exactly measures residual value sensitivity, and is computable through the feature-space products already used for aggregation. The matrices $G$ and $B$ are different products sharing centered keys; associativity is retained. This integration links entropy-derived calibration, current-key-dependent statistics, and residual sensitivity within one linear operator.

The conditioning argument explains query-wise scale calibration. For matched parameters and $0<z_{\min}<z_{\max}$, the linear map gives $\chi_{\mathrm{lin}}=(\lambda+z_{\min}/\alpha)/(\lambda+z_{\max}/\alpha)<\chi_\tau$, compressing the scale-span term in the bound more strongly. The rational rule balances calibration and amplitude preservation: its positive floor prevents unbounded gain and retains a vanishing residual as $x_i\to0$, its middle range partially calibrates scales, and its finite ceiling preserves variation among strong residuals. The mapping choice is supported by the ablations in Section~\ref{sec:temperature_ablations}.
\subsection{Complexity and Implementation}

For the large-axis case $M>d$, computing $G$, $B$, and the query-wise energies costs $O(Md^2)$ time. The shared matrices occupy $O(d^2)$ storage, token activations and intermediates occupy $O(Md)$, and temperatures occupy $O(M)$. Vanilla softmax attention requires $O(M^2d)$ time and $O(M^2)$ attention memory. Table~\ref{tab:complexity} focuses on this large-axis case, because it provides the main efficiency benefit.

For a short attention axis, the implementation evaluates the output directly as $(Q\widehat K^\top)V$. The squared-score energy is still computed from $\widehat K^\top\widehat K$, which is algebraically identical to summing the squared entries of $Q\widehat K^\top$. This choice does not change the method definition or the output. It only affects the order of intermediate multiplications for more efficient computation. In our setting, the temporal axis is short, so most of the measured saving comes from the spatial branch.

\begin{table}[t]
\centering
\caption{Large-axis complexity per head. Shared matrix state excludes $O(Md)$ token activations/intermediates and the proposed $O(M)$ temperatures; it is not total training memory.}
\label{tab:complexity}
\footnotesize
\begin{tabular}{@{}lcc@{}}
\toprule
Module & Time & Shared matrix state \\
\midrule
Softmax attention & $O(M^2d)$ & $O(M^2)$ \\
Proposed ($M>d$) & $O(Md^2)$ & $O(d^2)$ \\
\bottomrule
\end{tabular}
\end{table}

\begin{algorithm}[t]
\caption{Entropy-Calibrated Linear Attention (ECLA, large-axis)}
\label{alg:ecla}
\begin{algorithmic}[1]
\REQUIRE $Q,K,V\in\mathbb{R}^{M\times d}$, $\lambda>0$, $\alpha>0$ and $\beta>0$
\ENSURE $O\in\mathbb{R}^{M\times d}$
\STATE $\widehat K\gets K-M^{-1}\mathbf1\mathbf1^\top K$
\STATE $G\gets\widehat K^\top\widehat K$, $B\gets\widehat K^\top V$, and $\bar v\gets M^{-1}\mathbf1^\top V$
\FOR{$i=1$ to $M$}
    \STATE $x_i\gets q_iGq_i^\top/d$
    \STATE $z_i\gets x_i/(M\log M)$
    \STATE $\tau_i\gets\beta\sg\!\left(\sqrt{\lambda+z_i/(z_i+\alpha)}\right)$
    \STATE $o_i\gets\bar v+(q_iB)/(M\sqrt d\,\tau_i)$
\ENDFOR
\RETURN $O=[o_1^\top,\ldots,o_M^\top]^\top$
\end{algorithmic}
\end{algorithm}

ELinFormer follows the STAEformer~\cite{STAEformer} block design and replaces its temporal and spatial attention modules with the proposed operator. The embeddings, feed-forward networks, normalization layers, and prediction head are otherwise unchanged, enabling a controlled comparison of the attention module itself.

\section{Experiments}
\label{sec:experiments}

\subsection{Experimental Settings}

\textbf{Datasets.}
We evaluate the proposed method on five public multivariate benchmarks: PEMS-BAY, PEMS03, PEMS04, PEMS07, and PEMS08~\cite{pems, DCRNN}. These datasets are collected from the Caltrans Performance Measurement System and are widely used to evaluate spatio-temporal forecasting models. Dataset statistics are summarized in Table~\ref{tab:datasets}.

\begin{table}[t]
    \centering
\caption{Dataset statistics.}
    \label{tab:datasets}
    \small
    \setlength{\tabcolsep}{4pt}
    \begin{tabular}{lcccc}
    \toprule
        Dataset & \#Nodes & \#Edges & \#Timesteps & Interval \\ \midrule
        PEMS-BAY & 325 & 2369 & 52128 & 5 min \\
        PEMS03 & 358 & -- & 26208 & 5 min \\
        PEMS04 & 307 & 340 & 16992 & 5 min \\
        PEMS07 & 883 & 866 & 28224 & 5 min  \\
        PEMS08 & 170 & 295 & 17856 & 5 min \\
    \bottomrule
    \end{tabular}
\end{table}

\begin{table*}[t]
\centering
\caption{Overall forecasting results on highway traffic datasets. Best results are in bold and second-best results are underlined. Some unavailable entries are omitted. $^\dagger$ TSNN results are taken directly from the original paper and may not be fully comparable because they are not obtained from the common benchmark implementation.}
\label{tab:main_results}
\small
\begin{tabular}{l|ll|ll|ll|ll|ll}
\toprule
\multirow{2}{*}{Model} & \multicolumn{2}{c|}{PEMS-BAY} & \multicolumn{2}{c|}{PEMS03} & \multicolumn{2}{c|}{PEMS04} & \multicolumn{2}{c|}{PEMS07} & \multicolumn{2}{c}{PEMS08} \\
& MAE & RMSE & MAE & RMSE & MAE & RMSE & MAE & RMSE & MAE & RMSE \\
\midrule
STID~\cite{stid} & -- & -- & 15.37 & 25.10 & 18.68 & 30.61 & 20.07 & 33.39 & 14.55 & 23.85\\
M3-Net~\cite{M3_Net} & -- & -- & 14.88 & 25.49 & 18.30 & 30.15 & 19.42 & 32.85 & 13.89 & 23.33\\
TSNN$^\dagger$~\cite{tsnn} & -- & -- & 15.21 & \textbf{24.44} & 19.28 & 31.69 & 20.37 & 34.56 & 14.57 & 24.80\\
\midrule
AGCRN~\cite{AGCRN} & 1.60 & 3.67 & 15.24 & 26.65 & 19.28 & 31.02 & 20.68 & 34.45 & 15.78 & 24.76 \\
DCRNN~\cite{DCRNN} & 1.59 & 3.69 & 15.54 & 27.18 & 19.66 & 31.18 & 21.16 & 34.15 & 15.23 & 24.17 \\
GTS~\cite{GTS} & 1.68 & 3.79 & 15.41 & 26.15 & 21.32 & 33.55 & 22.47 & 35.42 & 16.92 & 26.68 \\
D2STGNN~\cite{D2STGNN} & \textbf{1.52} & \underline{3.53} & \underline{14.63} & 26.31 & 18.32 & \underline{29.89} & 19.49 & \underline{32.59} & 14.10 & 23.36 \\
HimNet~\cite{HimNet} & -- & -- & 15.11 & 26.56 & \textbf{18.14} & \textbf{29.88} & 19.21 & 32.75 & 13.57 & \textbf{23.22} \\
\midrule
STGCN~\cite{STGCN} & 1.63 & 3.72 & 15.83 & 27.51 & 19.76 & 31.51 & 22.25 & 35.83 & 16.19 & 25.51 \\
MTGNN~\cite{MTGNN} & 1.60 & 3.71 & 14.85 & 25.23 & 19.13 & 31.03 & 21.01 & 34.14 & 15.25 & 24.22 \\
GWNet~\cite{GWNet} & 1.59 & 3.68 & \textbf{14.59} & 25.24 & 18.53 & 29.92 & 20.44 & 33.38 & 14.67 & 23.55 \\
\midrule
PDFormer~\cite{PDformer} & 1.57 & 3.58 & 14.94 & 25.39 & 18.36 & 30.03 & 19.83 & 32.87 & 13.58 & 23.41 \\
STWave~\cite{STWave} & -- & -- & 14.93 & 26.50 & 18.50 & 30.39 & 19.94 & 33.88 & \underline{13.42} & 23.40 \\
RIPCN~\cite{ripcn} & -- & -- & 15.28 & 26.05 & 19.03 & 30.19 & -- & -- & 15.14 & 23.76\\
STAEformer~\cite{STAEformer} & \underline{1.54} & \textbf{3.52} & 15.35 & 27.55 & \underline{18.22} & 30.18 & \underline{19.14} & 32.60 & 13.46 & 23.25 \\
ELinFormer (Ours) & \underline{1.54} & 3.58 & 14.81 & \underline{24.86} & \textbf{18.14} & \textbf{29.88} & \textbf{19.06} & \textbf{32.39} & \textbf{13.34} & \underline{23.23} \\
\bottomrule
\end{tabular}
\end{table*}

\textbf{Baselines.}
We use STAEformer as the main vanilla-attention baseline because it is an attention-only model with strong performance in time series forecasting. It therefore enables a controlled comparison, where ELinFormer keeps the same backbone and replaces only the attention operator. We additionally compare with Linformer~\cite{linformer}, a low-rank attention method, and NaLaFormer~\cite{nalaformer}, a norm-aware linear attention method, in the same backbone. These comparisons evaluate efficient aggregation and norm-dependent selectivity alongside the proposed calibration. We also compare against representative recurrent, convolutional, graph-based, and Transformer-based forecasting models, including D2STGNN~\cite{D2STGNN}, HimNet~\cite{HimNet}, MTGNN~\cite{MTGNN}, PDFormer~\cite{PDformer}, STWave~\cite{STWave} and other baselines listed in Table~\ref{tab:main_results}. These broader baselines mainly provide a performance comparison, while the direct replacement target is STAEformer.

\textbf{Implementation details.}
We follow the preprocessing pipeline and evaluation protocol of BasicTS~\cite{exploring}. Experiments use PyTorch 1.10.0, Python 3.8, and CUDA 11.3 on an NVIDIA RTX 3080 GPU. Detailed hardware specifications and software dependencies are provided in the supplementary material. The data split, model width, optimizer, and other training hyperparameters follow the vanilla STAEformer configuration unless otherwise stated. We set $\lambda=10^{-2}$ and $\beta=1$. We evaluate $\alpha\in\{0.50,1.00,1.33,1.50,2.00\}$ on the
PEMS08 validation split and select $\alpha=1.33$ based on validation MAE. To avoid dataset-specific tuning, the selected value is fixed for all remaining datasets, heads, and queries. The query-dependent temperature is detached as shown in Algorithm~\ref{alg:ecla}. The KL histogram in Figure~\ref{fig:framework}(a) is computed offline from trained vanilla STAEformer attention on PEMS08 and does not enter the ELinFormer training path. Unless otherwise stated, results produced by our implementation are obtained from a single training run with the fixed data split and random seed, and peak GPU memory and average seconds per epoch are measured under the same implementation settings.

\textbf{Evaluation metrics.}
We report Mean Absolute Error (MAE), Root Mean Squared Error (RMSE), and Mean Absolute Percentage Error (MAPE). MAE measures average absolute error, RMSE places greater weight on large errors, and MAPE reports relative error. Missing or invalid entries are masked before metric computation.

\subsection{Comparison with Forecasting Baselines}

Table~\ref{tab:main_results} compares ELinFormer with representative forecasting models, and ELinFormer achieves competitive performance across all benchmarks. It obtains the lowest reported MAE on PEMS07 and PEMS08, matches STAEformer in MAE on PEMS-BAY, and remains close to the strongest baselines on PEMS03 and PEMS04. These results indicate that replacing softmax attention with the linear operator may not require a substantial loss of forecasting accuracy.

The comparison also reflects differences among modeling assumptions. Several strong graph-based methods exploit a predefined topology, a learned adjacency matrix, or a sparse dependency structure. In contrast, ELinFormer models dynamic cross-variable dependencies without requiring a predefined graph, a learned sparse mask, or a task-specific selection threshold. It retains the embedding, feed-forward, normalization, and prediction components of the Transformer backbone and changes only the attention operator. The results therefore suggest that calibrated linear aggregation can preserve useful dynamic dependency modeling while avoiding the dense pairwise attention matrix.

Performance nevertheless depends on the dataset and metric. Graph-based models remain particularly strong on PEMS-BAY, where the road topology may provide an informative structural prior. On PEMS03, ELinFormer improves over the vanilla-attention backbone but does not achieve the lowest MAE among all methods, whereas on PEMS07 and PEMS08 it performs favorably despite using no explicit graph. The difference between MAE and RMSE rankings further suggests that models may behave differently on typical errors and larger deviations. Accordingly, ELinFormer should be viewed as an efficient and broadly applicable dynamic dependency module rather than a universally superior forecasting model. The controlled comparisons in Table~\ref{tab:attention_compare} further isolate the accuracy and efficiency effects of replacing the attention operator.

\subsection{Controlled Accuracy and Efficiency Comparison}

\begin{table*}[t]
\centering
\caption{Controlled comparison in the same STAEformer backbone. Parenthesized values report relative changes from STAEformer. OOM denotes an out-of-memory failure, and ``--'' denotes an unavailable result.}
\label{tab:attention_compare}
\small
\setlength{\tabcolsep}{4.2pt}
\newcommand{\chg}[1]{\,{\footnotesize(#1)}}
\begin{tabular}{c|c|ccc|cc}
\toprule
Dataset & Model & MAE & RMSE & MAPE
& Memory (MiB) & Time (s/epoch) \\
\midrule
\multirow{4}{*}{PEMS03}
& STAEformer
& 15.35 & 27.55 & \underline{15.18\%}
& 8336 & 186 \\
& Linformer
& 15.12 & \underline{26.06} & 15.60\%
& \underline{5611} & \textbf{140} \\
& NaLaFormer
& \textbf{14.75} & 26.13 & \textbf{15.00\%}
& 7155 & 157 \\
& ELinFormer
& \underline{14.81}\chg{$\downarrow$3.5\%}
& \textbf{24.86}\chg{$\downarrow$9.8\%}
& 15.21\%\chg{$\uparrow$0.2\%}
& \textbf{5278}\chg{$\downarrow$36.7\%}
& \underline{151}\chg{$\downarrow$18.8\%} \\
\midrule
\multirow{4}{*}{PEMS04}
& STAEformer
& \underline{18.22} & 30.18 & \underline{11.98\%}
& 7096 & 101 \\
& Linformer
& 18.39 & 30.15 & 12.10\%
& \underline{5217} & \textbf{79} \\
& NaLaFormer
& 18.24 & \underline{29.97} & 12.03\%
& 5981 & 89 \\
& ELinFormer
& \textbf{18.14}\chg{$\downarrow$0.4\%}
& \textbf{29.88}\chg{$\downarrow$1.0\%}
& \textbf{11.92\%}\chg{$\downarrow$0.5\%}
& \textbf{5098}\chg{$\downarrow$28.2\%}
& \underline{86}\chg{$\downarrow$14.9\%} \\
\midrule
\multirow{4}{*}{PEMS07}
& STAEformer
& -- & -- & -- & OOM & -- \\
& Linformer
& -- & -- & -- & OOM & -- \\
& NaLaFormer
& -- & -- & -- & OOM & -- \\
& ELinFormer
& \textbf{19.06} & \textbf{32.39} & \textbf{8.04\%}
& \textbf{9948} & \textbf{375} \\
\midrule
\multirow{4}{*}{PEMS08}
& STAEformer
& \underline{13.46} & 23.25 & \underline{8.88\%}
& 4470 & 59 \\
& Linformer
& 13.50 & \textbf{23.22} & 8.95\%
& \underline{3933} & \textbf{55} \\
& NaLaFormer
& 13.47 & 23.37 & 8.90\%
& 4311 & \underline{56} \\
& ELinFormer
& \textbf{13.34}\chg{$\downarrow$0.9\%}
& \underline{23.23}\chg{$\downarrow$0.1\%}
& \textbf{8.78\%}\chg{$\downarrow$1.1\%}
& \textbf{3834}\chg{$\downarrow$14.2\%}
& \textbf{55}\chg{$\downarrow$6.8\%} \\
\bottomrule
\end{tabular}
\end{table*}

Table~\ref{tab:attention_compare} compares the proposed module with vanilla STAEformer, Linformer~\cite{linformer}, and NaLaFormer~\cite{nalaformer} in the same backbone. Linformer uses projection dimension $k=64$ across all datasets and layers. NaLaFormer adds a direct comparison with norm-aware linear attention. For STAEformer, PEMS07 accuracy in Table~\ref{tab:main_results} is benchmark-reported, whereas its controlled run in this table reached OOM.

Relative to STAEformer, ELinFormer reduces MAE by $0.4$--$3.5\%$ on the completed datasets, peak GPU memory by $14.2$--$36.7\%$, and per-epoch time by $6.8$--$18.8\%$. Relative to Linformer, it achieves lower MAE and peak memory on every completed dataset; Linformer is faster on PEMS03/04 and has marginally lower PEMS08 RMSE. Relative to NaLaFormer, ELinFormer has lower RMSE, memory, and epoch time on all three completed datasets, and lower MAE/MAPE on PEMS04/08. NaLaFormer obtains the lowest PEMS03 MAE ($14.75$) and MAPE ($15.00\%$), while ELinFormer uses $26.2\%$ less memory there.

The end-to-end memory and runtime reductions do not scale exactly with the theoretical attention complexity. The embedding layers, feed-forward or MLP blocks, normalization, prediction head, and data movement remain unchanged and account for a larger fraction of the total cost on smaller datasets. Consequently, the practical gain is less pronounced when the attention matrix is not the dominant computational or memory bottleneck. On PEMS07, both vanilla attention and Linformer reach the memory limit under the reported setting,
whereas ELinFormer remains trainable. NaLaFormer also reaches OOM on PEMS07.

While Table~\ref{tab:main_results} compares forecasting models broadly, Table~\ref{tab:attention_compare} isolates vanilla, low-rank, norm-aware, and entropy-calibrated attention within the same backbone.

\subsection{Temperature-Rule Analysis on PEMS08}
\label{sec:temperature_ablations}

\begin{figure*}[t]
\centering
\includegraphics[width=\linewidth]{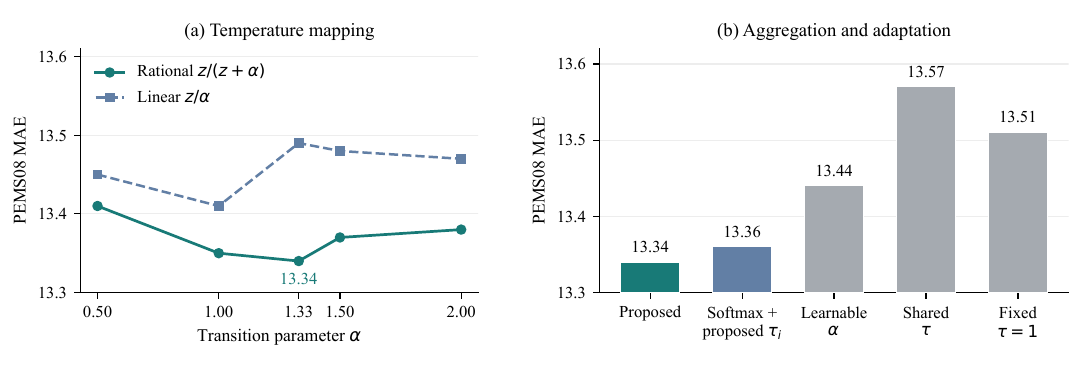}
\caption{PEMS08 single-run temperature analyses (lower MAE is better). (a) Rational and matched linear mappings over the same five $\alpha$ candidates under a common shortened schedule. (b) Five aggregation/adaptation variants, including dense softmax with the proposed temperature. ``Shared $\tau$'' is a single global mean computed on the validation split at the best $\alpha$. In the learnable-$\alpha$ ablation, gradients are stopped through $z_i$ and retained with respect to $\alpha$. The MAE axes display restricted ranges; exact values are listed in Supplement Tables~S4 and~S5.}
\label{fig:temperature_analysis}
\end{figure*}

To better isolate the role of the temperature rule, we perform a focused analysis on PEMS08. Figure~\ref{fig:temperature_analysis} reports two complementary observations. The left panel compares the proposed rational mapping $r(z;\alpha)=z/(z+\alpha)$ with the linear mapping $r(z;\alpha)=z/\alpha$ under several values of $\alpha$. The rational mapping achieves lower MAE than the linear mapping for all matched values of $\alpha$. The right panel compares five aggregation and temperature-adaptation variants, including dense softmax with the proposed temperature. Replacing the proposed query-specific rule with a learnable $\alpha$, a shared temperature, or a fixed temperature $\tau=1$ yields higher MAE in this analysis.

These results provide evidence for two design choices. First, the function shape matters: a bounded rational mapping is more effective in these runs than a linear rule that asymptotically forces high-energy queries toward a shared KL level. Second, the gain does not come from merely tuning a global temperature scalar. The proposed improvement depends on combining the bounded mapping with the query-specific score-energy statistic.

The proposed linear operator obtains $13.34$ MAE, dense softmax with the same query-wise temperature $13.36$, learnable $\alpha$ $13.44$, shared global temperature $13.57$, and fixed $\tau=1$ $13.51$. The softmax variant evaluates the temperature policy with exponential aggregation, while the shared and fixed scales examine the role of query-wise adaptation. These controls connect the entropy and selectivity questions raised by NaLaFormer and PolaFormer to the calibration choices in ELinFormer.

\subsection{Score-Energy Diagnostic on PEMS08}
\label{sec:score_energy_diagnostic}

\begin{figure*}[t]
\centering
\includegraphics[width=\linewidth,height=2.18in,keepaspectratio]{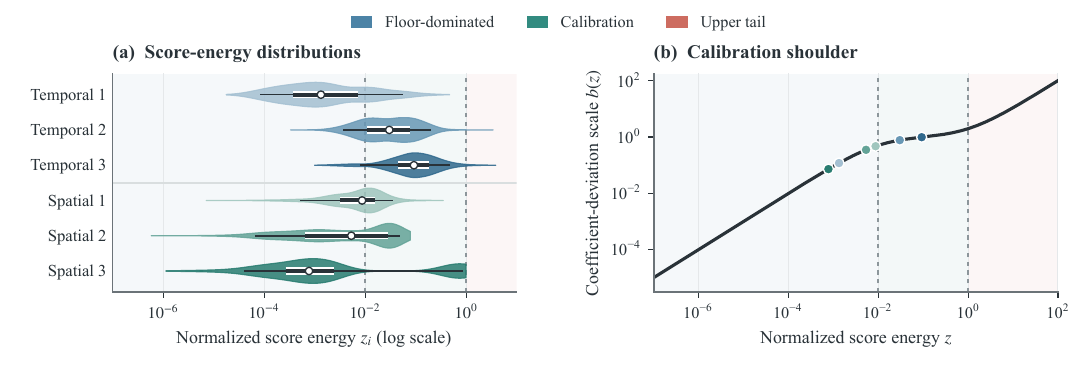}
\caption{PEMS08 score-energy diagnostic. (a) Distributions of $z_i=x_i/(M\log M)$ from $10{,}000$ archived samples per layer: temporal $M=12$, spatial $M=170$. Thin spans indicate the 5th--95th percentiles, thick spans the interquartile range, and white circles the medians. (b) Analytical coefficient-deviation scale $b(z)=z/\tau^2(z)$ on logarithmic axes; colored circles repeat the six layer medians from (a), showing where the sampled density is centered on the calibration curve. Shading and dashed guides indicate descriptive ranges separated at $\lambda\alpha/(1+\lambda)$ and $\alpha$. The diagnostic uses $\alpha=1$, $\lambda=10^{-2}$, and $\beta=1$; the selected forecasting configuration remains $\alpha=1.33$.}
\label{fig:score_energy_diagnostic}
\end{figure*}

Figure~\ref{fig:score_energy_diagnostic}(a) shows both layer dependence and within-layer spread. Temporal medians rise from $1.33\!\times\!10^{-3}$ to $9.23\!\times\!10^{-2}$, while spatial layer 3 combines a low-energy majority with a separated higher-energy component. This spread motivates a query-specific statistic rather than one scale per layer. At $\alpha=1$, $49.10\%$ of samples are floor-dominated, $50.70\%$ lie in the calibration range, and $0.19\%$ are in the upper tail.

Figure~\ref{fig:score_energy_diagnostic}(b) links these distributions to the operator through $M\|\widetilde{\mathbf a}_i-u\|_2^2=\log M\,b(z_i)$ in Eq.~\eqref{eq:linear_kl_link}. The curve gives small low-energy deviations, a slower-growth calibration range, and stronger high-energy growth. Supplement Section~S8 provides the temperature curve, occupancy, and reproduction details.

\section{Discussion and Limitations}
\label{sec:limitations}

The KL proxy is derived from a second-order expansion around uniform softmax attention and may be less accurate for highly concentrated score distributions. Its local entropy interpretation applies when the normalized score contrasts $s_i/\tau_i$ are small. For the deployed linear operator, the same proxy has the exact interpretation in Eq.~\eqref{eq:linear_kl_link}, controlling the squared deviation of its signed coefficients from uniform mixing.

\textbf{Interpretation of the high-energy regime.}
The normalized score energy also indicates when the local entropy interpretation should be examined more closely. Since
\begin{equation}
\sum_{j=1}^{M}s_{ij}=0,
\qquad
\sum_{j=1}^{M}s_{ij}^{2}=M\log M\,z_i,
\end{equation}
the normalized score vector satisfies
\begin{equation}
\left\|\frac{s_i}{\tau_i}\right\|_{\infty}
\geq\frac{\|s_i\|_2}{\sqrt{M}\tau_i}
=\frac{\sqrt{z_i\log M}}{\tau_i}.
\label{eq:high_energy_score_bound}
\end{equation}
In particular, in the upper-energy regime $z_i\geq z_{\mathrm{high}}=\alpha$, the bounded temperature rule gives
\begin{equation}
\left\|\frac{s_i}{\tau_i}\right\|_{\infty}
\geq\frac{1}{\beta}\sqrt{\frac{\alpha\log M}{1+\lambda}},
\label{eq:high_energy_temperature_bound}
\end{equation}
because $\tau_i^2<\beta^2(1+\lambda)$. All experiments use $\beta=1$. As $z_i$ grows, the lower bound in Eq.~\eqref{eq:high_energy_score_bound} also grows, so at least one normalized score must have substantial magnitude. The resulting softmax concentration additionally depends on the higher-order structure of the score vector.

For a realized score vector, concentration can be assessed directly using
\begin{equation}
\begin{aligned}
p_{ij}(\tau_i)&=\frac{\exp(s_{ij}/\tau_i)}
{\sum_{\ell=1}^{M}\exp(s_{i\ell}/\tau_i)},\\
H_i&=-\sum_{j=1}^{M}p_{ij}(\tau_i)\log p_{ij}(\tau_i).
\end{aligned}
\label{eq:exact_high_energy_softmax}
\end{equation}
If $p_{i,\max}\geq\rho$ for $\rho\in[1/M,1]$, then
\begin{equation}
H_i\leq h_2(\rho)+(1-\rho)\log(M-1),
\label{eq:entropy_concentration_bound}
\end{equation}
where $h_2(\rho)=-\rho\log\rho-(1-\rho)\log(1-\rho)$ is binary entropy. This bound quantifies entropy reduction for sufficiently concentrated queries. Supplement Section~S12 gives its proof and relates the local Taylor condition to score contrast.

The analyzed score-energy sample predominantly occupies the low- and moderate-energy regions; at the diagnostic reference $\alpha=1$, the upper tail accounts for $0.19\%$ of sampled values. Within these regions, the expansion provides a local entropy interpretation when the normalized contrasts are small. Energy alone does not determine $p_{i,\max}$. In the upper-energy region, exact softmax entropy in Eq.~\eqref{eq:exact_high_energy_softmax} provides the direct concentration diagnostic. True softmax KL is bounded by $\log M$ and entropy lies in $[0,\log M]$; the continued growth of the second-order proxy describes the linear operator's deviation scale beyond that local approximation.

The deployed signed operator retains a monotonic interpretation throughout this regime. From Eq.~\eqref{eq:linear_kl_link},
\begin{equation}
M\left\|\widetilde{\mathbf a}_i-u\right\|_2^2
=\frac{x_i}{M\tau_i^2}
=2\widehat{\mathrm{KL}}_i,
\label{eq:high_energy_signed_deviation}
\end{equation}
which is strictly increasing with $z_i$ by Eq.~\eqref{eq:b_derivative}. High-energy queries therefore retain a larger signed deviation from uniform aggregation, preserving their score-energy ordering and capacity for query-dependent correction. This is a correlation-based calibration of signed coefficients; the realized output also depends on the values being aggregated. The exact entropy calculation supplies the corresponding softmax concentration measure.

\textbf{Scope and further evaluation.}
The operator analysis fixes queries and keys and concerns projected value activations on the stated row subsystem. Extending it to coupled query/key learning and full-network optimization is a further theoretical direction. The current evaluation focuses on PEMS forecasting benchmarks, with $\lambda$ fixed and $\alpha$ selected on PEMS08 validation data. Additional multivariate domains, repeated seeds, longer attention axes, and PolaFormer comparisons under the same backbone and training budget would broaden the assessment of robustness, selectivity, and scalability. Fused implementations of the shared products would also help evaluate the attainable runtime gains.

\section{Conclusion}

We proposed Entropy-Calibrated Linear Attention for efficient multivariate time series modeling. The method uses a second-order KL proxy to connect centered score energy, query-specific temperature, and estimated attention entropy. A smooth temperature rule retains vanishing low-energy residuals without unbounded gain, approximately aligns moderate-energy queries, and avoids a fixed KL ceiling for high-energy queries. Shared centered-key products compute both the proxy and the linear attention output without forming the large pairwise attention matrix. Experiments on five PEMS benchmarks show competitive accuracy together with lower memory usage and training time than vanilla attention, and a favorable accuracy--memory trade-off relative to Linformer.

\IEEEtriggeratref{17}
\bibliographystyle{IEEEtran}
\bibliography{elinformer}
\end{document}


\maketitle

\textbf{Code completeness.} The Code Supplement includes preprocessing, training, evaluation, controlled STAEformer comparisons, temperature ablations, and figure generation. Raw datasets and checkpoints are omitted owing to the file-size limit; reproduction requires the specified datasets, environment, and training configurations.

\section{Scope of the Supplement}
This supplement records the environment, data preparation, configurations, commands, PEMS08 temperature analyses, and score-energy diagnostic. Section~\ref{sec:supp_operator} proves the operator results, and Section~\ref{sec:supp_entropy_scope} gives the entropy bound and high-energy interpretation used in the discussion. The implementation uses archived BasicTS code; the main paper's broad comparison also includes literature or public-benchmark values.

\section{Computing Environment}
The locally reported experiments were conducted in the environment summarized in Table~\ref{tab:env}. Dependency records are supplied in both \code{requirements.txt} and \code{environment.yml}.

\begin{table}[t]
\centering
\caption{Computing environment used for the locally reported experiments.}
\label{tab:env}
\begin{tabular}{@{}lp{0.64\columnwidth}@{}}
\toprule
Component & Specification \\
\midrule
Operating system & Linux \\
CPU & 12 vCPU Intel Xeon Platinum 8255C, 2.50 GHz \\
System memory & 40 GB \\
GPU & NVIDIA RTX 3080, 10 GB VRAM \\
Python & 3.8 \\
PyTorch & 1.10.0 \\
CUDA / cuDNN & 11.3 / 8.2.0 \\
\bottomrule
\end{tabular}
\end{table}

\section{Datasets and Preprocessing}
We use the five public PEMS benchmarks listed in Table~\ref{tab:data}. They contain multivariate sensor measurements sampled at five-minute intervals. The exact train/validation/test construction, normalization, sliding windows, and masking protocol follow the included BasicTS pipeline. No private dataset is used.

\begin{table}[t]
\centering
\caption{Dataset statistics, matching the main paper. A dash denotes unavailable graph-edge information.}
\label{tab:data}
\begin{tabular}{lrrrr}
\toprule
Dataset & Nodes & Edges & Time steps & Interval \\
\midrule
PEMS-BAY & 325 & 2369 & 52128 & 5 min \\
PEMS03 & 358 & -- & 26208 & 5 min \\
PEMS04 & 307 & 340 & 16992 & 5 min \\
PEMS07 & 883 & 866 & 28224 & 5 min \\
PEMS08 & 170 & 295 & 17856 & 5 min \\
\bottomrule
\end{tabular}
\end{table}

Place raw datasets under \code{datasets/raw_data} and run
\begin{center}
\code{python scripts/data_preparation/DATASET/generate_training_data.py}
\end{center}
Some inherited utilities load Python pickle files because that is the format of several public benchmark artifacts. These files should be treated as trusted inputs.

\section{Final Configurations}
All active model and training hyperparameters are stored under \code{baselines/}. Table~\ref{tab:config} lists the method-specific settings that differ from or extend the STAEformer backbone.

\begin{table}[t]
\centering
\caption{Method-specific final configurations.}
\label{tab:config}
\begin{tabular}{p{0.25\linewidth}p{0.64\linewidth}}
\toprule
Setting & Final value or selection procedure \\
\midrule
Temperature floor & $\lambda=10^{-2}$, fixed \\
Global scale & $\beta=1$, fixed \\
$\alpha$ candidates & $\{0.50,1.00,1.33,1.50,2.00\}$ \\
Selected $\alpha$ & $1.33$, selected by PEMS08 validation MAE \\
Cross-dataset use & The selected $\alpha$ is reused for the remaining PEMS datasets, heads, and queries \\
Backbone settings & Follow the corresponding STAEformer configurations unless explicitly changed \\
\bottomrule
\end{tabular}
\end{table}

For raw centered squared-score energy $x_i$, attention-axis length $M$, and normalized energy $z_i=x_i/(M\log M)$, the proposed implementation uses
\begin{equation}
\begin{aligned}
\tau_i&=\beta\operatorname{sg}\!\left(\sqrt{\lambda+\frac{x_i}{\alpha M\log M+x_i}}\right)\\
      &=\beta\operatorname{sg}\!\left(\sqrt{\lambda+\frac{z_i}{z_i+\alpha}}\right),
\qquad \lambda=10^{-2},\quad\beta=1.
\end{aligned}
\label{eq:rational}
\end{equation}
The query-dependent temperature computation is detached, and $\beta=1$ is fixed. The matched linear temperature-map ablation uses the same $\lambda=10^{-2}$ and $\beta=1$:
\begin{equation}
\tau_i^{\mathrm{lin}}=\sqrt{10^{-2}+\frac{x_i}{\alpha M\log M}}
      =\sqrt{10^{-2}+\frac{z_i}{\alpha}}.
\label{eq:linear}
\end{equation}
Explicit PEMS08 candidate configurations cover all five reported values of $\alpha$. The focused function comparison uses a common shortened schedule across mappings.

The controlled comparison in main-paper Table~IV retains the STAEformer backbone for vanilla attention, Linformer~\cite{linformer}, NaLaFormer~\cite{nalaformer}, and ELinFormer. Linformer uses projection dimension $k=64$ across all datasets and layers; other backbone and training settings follow the corresponding STAEformer configuration. Peak GPU memory and average seconds per epoch are measured under the same implementation settings. On PEMS07, all three baseline runs reach OOM, so their accuracy and time entries are unavailable. The STAEformer PEMS07 accuracy in main-paper Table~III is benchmark-reported and is distinct from its failed controlled run.

\section{Randomness and Number of Runs}
All supplied experiment configurations use seed $1$. The seed is applied to Python, NumPy, PyTorch, and CUDA where supported by the training pipeline. Each locally produced value is a single-run result. Deterministic CUDA execution is not enabled, so minor numerical or run-to-run differences may remain. Literature-reported baseline values in the broad comparison are not counted as locally executed runs.

\section{Code Organization}
The relevant package structure is:
\begin{itemize}\setlength\itemsep{0pt}
\item \code{scripts/data_preparation/}: dataset preprocessing;
\item \code{baselines/STAEformEntro/}: proposed rational temperature rule and ELinFormer configurations;
\item \code{baselines/STAEformEntro2/}: matched linear temperature-map ablations;
\item \code{baselines/STAEformer/}: controlled vanilla-attention comparison;
\item \code{experiments/}: training and inference entry points;
\item \code{figures/plot.py}: temperature-rule sensitivity and adaptation figure;
\item \code{figures/plot_hist.py}: PEMS08 score-energy diagnostic;
\item \code{figures/sumx2.npy}: six-layer numeric score-energy sample;
\item \code{BasicTS_requirements/}: archived BasicTS documentation, dependencies, and license.
\end{itemize}

\section{Reproduction Commands}
\subsection{Preprocess a dataset}
\code{python scripts/data_preparation/DATASET/generate_training_data.py}

\subsection{Train and evaluate ELinFormer}
\code{python experiments/train.py -c baselines/STAEformEntro/DATASET.py}\\
\code{python experiments/inference.py -c baselines/STAEformEntro/DATASET.py}

\subsection{Run the PEMS08 linear temperature-map ablation}
\code{python experiments/train.py -c baselines/STAEformEntro2/PEMS08.py}\\
\code{python experiments/inference.py -c baselines/STAEformEntro2/PEMS08.py}

\subsection{Generate analysis figures}
\code{python figures/plot.py}\\
\code{python figures/plot_hist.py}

The figure scripts resolve their default inputs and outputs relative to \code{figures/}. Configuration files remain the authoritative record of model and optimization settings.

\section{Additional Temperature Analysis}
\label{sec:supp_temperature}
All values in this section are single-run PEMS08 MAE results and are not accompanied by statistical significance tests.

\subsection{Exact Function-Sensitivity Values}
The proposed bounded rational rule is $r_{\mathrm{rat}}(z;\alpha)=z/(z+\alpha)$, whereas the matched linear alternative is $r_{\mathrm{lin}}(z;\alpha)=z/\alpha$. Table~\ref{tab:sens} reports the exact values visualized in main-paper Figure~2.

\begin{table}[t]
\centering
\caption{PEMS08 MAE for the rational and linear temperature mappings. Lower is better.}
\label{tab:sens}
\begin{tabular}{ccc}
\toprule
$\alpha$ & Rational $z/(z+\alpha)$ & Linear $z/\alpha$ \\
\midrule
0.50 & 13.41 & 13.45 \\
1.00 & 13.35 & 13.41 \\
1.33 & 13.34 & 13.49 \\
1.50 & 13.37 & 13.48 \\
2.00 & 13.38 & 13.47 \\
\bottomrule
\end{tabular}
\end{table}

\subsection{Additional Softmax-Temperature Experiment}
To separate the effect of the query-specific temperature from the linear aggregation rule, we additionally apply the same temperature directly to dense softmax attention,
\begin{equation}
p_{ij}=\frac{\exp(s_{ij}/\tau_i)}{\sum_{\ell=1}^{M}\exp(s_{i\ell}/\tau_i)}.
\end{equation}
This variant obtains PEMS08 MAE $13.36$, close to the $13.34$ MAE of ELinFormer. It tests the proposed query-wise policy in the dense softmax setting used by entropy and selectivity methods, while holding the policy itself fixed. This complements the mapping and temperature-adaptation comparisons in Table~\ref{tab:abl}.

\begin{table}[t]
\centering
\caption{Temperature-related PEMS08 ablations.}
\label{tab:abl}
\begin{tabular}{@{}p{0.80\columnwidth}r@{}}
\toprule
Variant & MAE \\
\midrule
Proposed linear operator, rational temperature & 13.34 \\
Softmax attention with the proposed $\tau_i$ & 13.36 \\
Learnable $\alpha$ & 13.44 \\
Shared global $\tau$ & 13.57 \\
Fixed $\tau=1$ & 13.51 \\
\bottomrule
\end{tabular}
\end{table}

``Shared global $\tau$'' uses a single global mean temperature computed on the validation split at the best $\alpha$. ``Fixed $\tau=1$'' removes adaptive temperature scaling. In the learnable-$\alpha$ ablation, gradients are stopped through $z_i$ but retained with respect to $\alpha$; this differs from the detached query-dependent temperature used by the final model.

\subsection{PEMS08 Score-Energy and Coefficient-Deviation Diagnostic}
The Code Supplement includes a six-layer, 10,000-value-per-layer sample of the unnormalized centered squared-score energy
\begin{equation}
x_i=\sum_{j=1}^{M}s_{ij}^{2}.
\end{equation}
The first three layers are temporal attention with $M=12$, and the final three are spatial attention with $M=170$. The diagnostic normalizes $x_i$ as $z_i=x_i/(M\log M)$ and recomputes $\tau_i$ from the same sampled values. This avoids an invalid element-wise comparison with a separately and independently sampled temperature array.

\begin{figure*}[t]
\centering
\includegraphics[width=\textwidth]{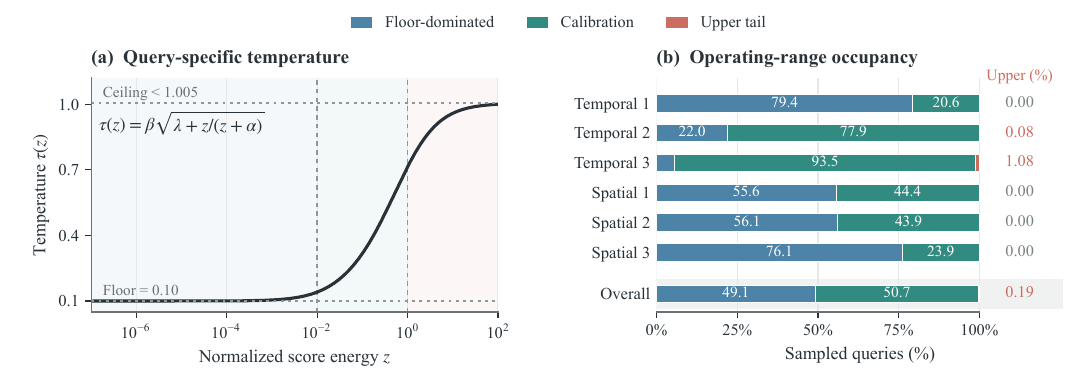}
\caption{Additional PEMS08 diagnostic at $\lambda=10^{-2}$, $\beta=1$, and reference $\alpha=1$. (a) Bounded analytical temperature rule. (b) Descriptive occupancy of the floor-dominated, calibration, and upper-tail ranges; the right column reports upper-tail percentages. Each layer contains $10{,}000$ archived samples. The boundaries $z_{\mathrm{floor}}=\lambda\alpha/(1+\lambda)$ and $z=\alpha$ are visualization guides rather than hard model gates. The selected forecasting configuration remains $\alpha=1.33$.}
\label{fig:diag}
\end{figure*}

The selected cross-dataset configuration remains $\alpha=1.33$. Main-paper Figure~3 and Figure~\ref{fig:diag} use the searched candidate $\alpha=1$ as an analytical reference, placing $r(\alpha)=1/2$ at $z=1$. Its PEMS08 MAE is $13.35$, versus $13.34$ at $\alpha=1.33$; the diagnostic leaves the selected configuration unchanged.

Main-paper Figure~3(a) shows substantial layer dependence on a common normalized energy axis. Figure~\ref{fig:diag}(a) supplies the corresponding bounded temperature rule. The layer-wise distributions span different parts of this curve, with within-layer variation inducing different query-specific temperatures.

Main-paper Figure~3(b) uses logarithmic axes for both $z$ and $b(z)$ so that multiplicative power laws appear as lines and the moderate-energy shoulder is visible. Its local log-log slope is
\begin{equation}
\frac{d\log b(z)}{d\log z}
=1-\frac{\alpha z}{(z+\alpha)^2[\lambda+z/(z+\alpha)]}.
\end{equation}
For $\lambda=10^{-2}$ and $\alpha=1$, the minimum is approximately $0.18$ near $z\simeq0.10$. This is a property of the analytical curve, not an experimental metric or threshold: the middle behavior is an approximate calibration shoulder rather than a hard flat plateau. The plotted quantity satisfies
\begin{equation}
\begin{aligned}
\widehat{\mathrm{KL}}_i&=\frac{\log M}{2}b(z_i),\\
M\lVert\widetilde{\mathbf a}_i-\mathbf u\rVert_2^2&=\log M\,b(z_i).
\end{aligned}
\end{equation}
where the second identity is exact for the deployed signed linear coefficients.

Figure~\ref{fig:diag}(b) uses $z_{\mathrm{floor}}=\lambda\alpha/(1+\lambda)\simeq0.0099$ and $z=\alpha=1$ as descriptive boundaries. Table~\ref{tab:occ} reports the corresponding occupancy and median temperature.

\begin{table*}[t]
\centering
\caption{Descriptive occupancy of the PEMS08 diagnostic ranges. Each layer is a 10,000-value sample with replacement.}
\label{tab:occ}
\begin{tabular}{lrrrr}
\toprule
Layer & Floor (\%) & Transition (\%) & Upper tail (\%) & Median $\tau$ \\
\midrule
Temporal 1 & 79.38 & 20.62 & 0.00 & 0.1064 \\
Temporal 2 & 21.99 & 77.93 & 0.08 & 0.1977 \\
Temporal 3 & 5.47 & 93.45 & 1.08 & 0.3074 \\
Spatial 1 & 55.59 & 44.41 & 0.00 & 0.1364 \\
Spatial 2 & 56.10 & 43.90 & 0.00 & 0.1236 \\
Spatial 3 & 76.08 & 23.92 & 0.00 & 0.1038 \\
\midrule
Overall & 49.10 & 50.70 & 0.19 & -- \\
\bottomrule
\end{tabular}
\end{table*}

Approximately half of the sampled values lie in the transition/calibration range, including $77.93\%$ and $93.45\%$ in temporal layers 2 and 3; $0.19\%$ lie in the empirical upper tail. The observed occupancy describes how the sampled layers use the bounded temperature curve. Predictive effectiveness is evaluated separately through the mapping and temperature-adaptation ablations.

\subsection{Correspondence to Entropy and Selectivity Designs}
\label{sec:supp_literature_controls}
NaLaFormer couples the query norm with attention spikiness through norm-aware feature maps~\cite{nalaformer}; PolaFormer uses polarity-aware interactions and a learnable power map to reduce entropy~\cite{polaformer}. These designs motivate examining how concentration is controlled within a linear operator. The rational-versus-linear comparison in Table~\ref{tab:sens} tests the tradeoff between partial calibration and amplitude preservation. Table~\ref{tab:abl} tests query-wise adaptation against shared or fixed scales, and evaluates the proposed temperature in dense softmax. Each control retains the recorded PEMS08 setting and result. Together they assess the calibration choices of ELinFormer within its forecasting backbone.

\section{Generative AI Usage}
A large language model assisted with manuscript revisions, theoretical derivations, algebraic checks, and presentation of the existing PEMS08 sample. Forecasting results, raw samples, and model configurations were taken from the existing records. Model training and checkpoint selection were outside this assistance. The authors are responsible for checking the final text, equations, figures, and evidence.

\section{Checkpoints, Release, and Third-Party Code}
Checkpoints are omitted owing to the file-size limit. BasicTS may derive checkpoint directories from configuration hashes, requiring explicit checkpoint selection after configuration edits. The archived license is in \code{BasicTS_requirements/LICENSE}, and \code{THIRD_PARTY_NOTICES.md} documents third-party material. The STAEformer-derived backbone retains its original license obligations; licenses for author-owned additions cover only those additions. All local accuracy values are single runs. Reproduction requires the omitted data, environment, and checkpoints or retraining.

\section{Operator Sensitivity and Conditional Row Conditioning}
\label{sec:supp_operator}
This section proves main-paper Proposition~1 and expands the operator-level interpretation in Eq.~(21). Fix $Q,K$ and an attention-axis length $M>1$, let the head width be $d$, and use positive parameters $\alpha,\beta,\lambda$. With centered keys, define
\begin{equation}
\begin{aligned}
 S&=Q\widehat K^\top/\sqrt d,\\
 D_\tau&=\operatorname{diag}(1/\tau_i),\qquad \mathcal J_\tau=D_\tau S/M.
\end{aligned}
 \label{eq:supp_jacobian}
\end{equation}
For one projected value channel $v\in\mathbb R^M$, the residual is $r=\mathcal J_\tau v$. The row-wise sensitivity is therefore the exact identity
\begin{equation}
\begin{aligned}
 \left\|\frac{\partial r_i}{\partial v}\right\|_2^2
 &= [\mathcal J_\tau\mathcal J_\tau^\top]_{ii}\\
 &= \frac{x_i}{M^2\tau_i^2}=\frac{2\widehat{\mathrm{KL}}_i}{M},
 \qquad x_i=\|s_i\|_2^2.
\end{aligned}
 \label{eq:supp_exact_sensitivity}
\end{equation}
Equation~\eqref{eq:supp_exact_sensitivity} concerns projected value activations with fixed queries and keys, rather than the full $W_Q,W_K,W_V$ parameter Jacobian. The equality is algebraic; only the approximation of softmax KL uses the local Taylor expansion. The linear coefficients are signed, not probabilities.

The complete output contains the uniform context and the residual. Since centered scores satisfy $S\mathbf 1_M=0$, the two value derivatives are orthogonal, and hence
\begin{equation}
 \left\|\frac{\partial o_i}{\partial v}\right\|_2^2
 =\frac{1}{M}+\frac{x_i}{M^2\tau_i^2}.
 \label{eq:supp_full_output_sensitivity}
\end{equation}
This decomposition separates the global context sensitivity from the calibrated residual sensitivity.

\subsection{Conditional Gram-Matrix Bound}
Let $I$ contain $n\geq2$ linearly independent rows of $S$. For $i,j\in I$, define
\begin{equation}
\begin{aligned}
 C_{ij}&=\frac{\langle s_i,s_j\rangle}{\sqrt{x_i x_j}},\\
 J_0&=S_{I,:}/M,\qquad J_\tau=(D_\tau S)_{I,:}/M,\\
 \Gamma_0&=J_0J_0^\top,\qquad \Gamma_\tau=J_\tau J_\tau^\top.
\end{aligned}
 \label{eq:supp_gram_defs}
\end{equation}
The matrices $\Gamma_0$ and $\Gamma_\tau$ are output-space Gram matrices for the residual value Jacobian. They are not a full-network Hessian or NTK. Let $z_{\min}$ and $z_{\max}$ denote the minimum and maximum of $z_i=x_i/(M\log M)$ over $I$.

\begin{samepage}
\begin{proposition}[Conditional row-conditioning bound]
\label{prop:supp_row_conditioning}
With the definitions above and with $C$ positive definite,
\begin{equation}
\begin{aligned}
 \frac{\kappa(\Gamma_\tau)}{\kappa(\Gamma_0)}
 &\leq \kappa(C)\,\chi_\tau,\\
 \chi_\tau&=
 \frac{\lambda+z_{\min}/(z_{\min}+\alpha)}
      {\lambda+z_{\max}/(z_{\max}+\alpha)}.
\end{aligned}
 \label{eq:supp_conditioning_bound}
\end{equation}
In particular, $\kappa(C)\chi_\tau<1$ is a sufficient condition for strict improvement of this local value-space subsystem, not a necessary condition.
\end{proposition}
\end{samepage}

\begin{proof}
Let $R_0=\operatorname{diag}(\sqrt{x_i})_{i\in I}$ and $R_\tau=\operatorname{diag}(\sqrt{x_i}/\tau_i)_{i\in I}$. Direct multiplication gives
\begin{equation}
 \Gamma_0=R_0 C R_0/M^2,
 \qquad \Gamma_\tau=R_\tau C R_\tau/M^2.
 \label{eq:supp_factorization}
\end{equation}
Here $\kappa$ denotes the spectral condition number. Independence of the normalized rows makes $C$ positive definite. For a positive diagonal $R$ and a nonzero $y$, the Rayleigh quotient satisfies
\begin{equation}
 \lambda_{\min}(C)\|Ry\|_2^2
 \leq y^\top RCRy\leq\lambda_{\max}(C)\|Ry\|_2^2.
 \label{eq:supp_rayleigh}
\end{equation}
Dividing by $\|y\|_2^2$ and using the minimum and maximum diagonal entries of $R^2$ gives
\begin{equation}
\begin{aligned}
 \lambda_{\max}(RCR)&\leq \lambda_{\max}(C)\max_i R_{ii}^2,\\
 \lambda_{\min}(RCR)&\geq \lambda_{\min}(C)\min_i R_{ii}^2.
\end{aligned}
\label{eq:supp_rayleigh_extrema}
\end{equation}
Applying Eq.~\eqref{eq:supp_rayleigh} to $R_\tau$ yields
\begin{equation}
 \kappa(\Gamma_\tau)\leq\kappa(C)
 \frac{\max_{i\in I}x_i/\tau_i^2}{\min_{i\in I}x_i/\tau_i^2}.
 \label{eq:supp_tau_bound}
\end{equation}
For $\Gamma_0$, its diagonal entries are $x_i/M^2$. Since $\lambda_{\max}$ is at least the largest diagonal entry and $\lambda_{\min}$ is at most the smallest one,
\begin{equation}
 \kappa(\Gamma_0)\geq x_{\max}/x_{\min}.
 \label{eq:supp_unscaled_bound}
\end{equation}
The implemented temperature is
\begin{equation}
\begin{aligned}
 \tau(z)&=\beta\sqrt{\lambda+\frac{z}{z+\alpha}},\\
 b(z)&=\frac{z}{\tau^2(z)}
 =\frac{z}{\beta^2(\lambda+z/(z+\alpha))}.
\end{aligned}
 \label{eq:supp_b_def}
\end{equation}
Its derivative is
\begin{equation}
 b'(z)=
 \frac{(1+\lambda)z^2+2\lambda\alpha z+\lambda\alpha^2}
 {\beta^2((1+\lambda)z+\lambda\alpha)^2}>0.
 \label{eq:supp_b_derivative}
\end{equation}
Thus the extrema of $x_i/\tau_i^2$ occur at the same rows as the extrema of $x_i$, and
\begin{equation}
 \frac{\max_i x_i/\tau_i^2}{\min_i x_i/\tau_i^2}
 =\frac{x_{\max}}{x_{\min}}\chi_\tau.
 \label{eq:supp_span_identity}
\end{equation}
Combining Eqs.~\eqref{eq:supp_tau_bound}, \eqref{eq:supp_unscaled_bound}, and \eqref{eq:supp_span_identity} proves Eq.~\eqref{eq:supp_conditioning_bound}.\end{proof}

\subsection{Local Quadratic Interpretation}
For fixed $Q,K$, a value-space proxy $J\in\{J_0,J_\tau\}$, and any target $y\in\mathbb R^n$, consider
\begin{equation}
\begin{aligned}
 \ell(v)&=\tfrac12\|Jv-y\|_2^2,\\
 e_t&=Jv_t-y,\qquad \Gamma=JJ^\top.
\end{aligned}
 \label{eq:supp_proxy_loss}
\end{equation}
Gradient descent on this proxy gives $v_{t+1}=v_t-\eta J^\top e_t$ and hence $e_{t+1}=(I-\eta\Gamma)e_t$. Since $J$ has full row rank, $\Gamma$ is positive definite. The eigenvalues of $I-\eta\Gamma$ are $1-\eta\lambda_j(\Gamma)$. Balancing the extreme absolute eigenvalues gives each proxy problem its own optimal positive constant step size
\begin{equation}
 \eta=\frac{2}{\lambda_{\max}(\Gamma)+\lambda_{\min}(\Gamma)},
 \label{eq:supp_opt_step}
\end{equation}
The resulting spectral norm of $I-\eta\Gamma$ is $(\kappa(\Gamma)-1)/(\kappa(\Gamma)+1)$. Iterating the residual recurrence gives
\begin{equation}
 \|e_t\|_2\leq
 \left(\frac{\kappa(\Gamma)-1}{\kappa(\Gamma)+1}\right)^t
 \|e_0\|_2.
 \label{eq:supp_proxy_rate}
\end{equation}
The factor is target-independent and worst-case, at each proxy's own optimal step size. It does not compare a common learning rate or guarantee improvement for each target or step. Actual values are activations, not independently optimized parameters, and this proxy does not prove forecasting-loss or optimizer convergence. With fixed $Q,K$, $\tau$ is independent of $v$, so detach leaves Eq.~\eqref{eq:supp_exact_sensitivity} unchanged; learned query and key projections remain coupled.

A positive shared $\beta$ multiplies the residual Gram by a common factor and hence leaves its condition number unchanged. For the same $\alpha,\lambda,z_{\min},z_{\max}$ and $0<z_{\min}<z_{\max}$, the linear map has
\begin{equation}
 \chi_{\mathrm{lin}}=\frac{\lambda+z_{\min}/\alpha}
 {\lambda+z_{\max}/\alpha}<\chi_\tau.
 \label{eq:supp_linear_vs_rational}
\end{equation}
To verify the inequality, set $a=z_{\min}$, $b=z_{\max}$, and $f(z)=\lambda+z/(z+\alpha)$. Then
\begin{equation}
\begin{aligned}
h(z)&=\frac{\lambda+z/\alpha}{f(z)},\\
h'(z)&=\frac{z((1+\lambda)z+2\lambda\alpha)}
{\alpha((1+\lambda)z+\lambda\alpha)^2}>0.
\end{aligned}
\label{eq:supp_span_comparison_proof}
\end{equation}
Thus $\chi_{\mathrm{lin}}/\chi_\tau=h(a)/h(b)<1$. The inequality says that the linear map compresses this bound's scale span more strongly; it does not establish that its actual condition number or forecasting accuracy is better. The rational map retains a positive temperature floor, a bounded transition, and a finite ceiling, giving a calibration and amplitude-preservation tradeoff. Its mapping choice is supported by the single-run ablations in Section~\ref{sec:supp_temperature}.

The subset $I$ is selected only for the statement of the proposition; the code computes temperatures for all queries. The complete $S$ satisfies $\operatorname{rank}(S)\leq\min(d,M-1)$, so its full row Gram can be singular. The score-energy diagnostic in Section~\ref{sec:supp_temperature} does not measure $C$ or verify the sufficient condition in Proposition~\ref{prop:supp_row_conditioning}.

The controls in Section~\ref{sec:supp_literature_controls} evaluate the ELinFormer temperature policy. Main-paper Table~IV separately reports the NaLaFormer operator comparison; a corresponding PolaFormer evaluation remains a further direction. The sampled energy percentages are descriptive diagnostics rather than independent observations or significance tests.

\section{Local Entropy Interpretation and High-Energy Scores}
\label{sec:supp_entropy_scope}

\subsection{Normalized contrasts and the Taylor expansion}
Let $t_j=s_{ij}/\tau_i$ for a fixed query, so $\sum_j t_j=0$, and define $F(t)=\log(M^{-1}\sum_j e^{t_j})$. Then
\begin{equation}
\operatorname{KL}(\operatorname{softmax}(t)\|u)
=t^\top\nabla F(t)-F(t).
\end{equation}
At $t=0$, $F(0)=0$, $\nabla F(0)=u$, and $\nabla^2 F(0)=M^{-1}I-uu^\top$. On the centered subspace, Taylor expansion therefore yields
\begin{equation}
\operatorname{KL}(\operatorname{softmax}(t)\|u)
=\frac{\|t\|_2^2}{2M}+O(\|t\|_2^3),
\label{eq:supp_taylor_scope}
\end{equation}
for fixed $M$ as $t\to0$. This is the local basis for the main-paper KL proxy. The relevant small quantity is the temperature-normalized contrast, rather than $z_i$ alone. In particular, $\tau_i\geq\beta\sqrt\lambda$ implies
\begin{equation}
\|t\|_\infty\leq\frac{\sqrt{M\log M\,z_i}}{\beta\sqrt\lambda},
\end{equation}
so $z_i\to0$ gives the local limit at fixed $M,\beta,\lambda$. Finite low- or moderate-energy ranges provide descriptive operating regions; checking the local approximation at a realized query requires its normalized contrasts.

\subsection{Entropy bound from a maximum probability}
For any distribution $p$ on $M>1$ positions, let $q=\max_j p_j\geq1/M$. Grouping the largest entry and the remaining mass gives
\begin{equation}
H(p)=h_2(q)+(1-q)H(\bar p)
\leq h_2(q)+(1-q)\log(M-1),
\end{equation}
where $\bar p$ is the distribution of the remaining entries normalized by $1-q$; at $q=1$ the expression is interpreted by continuity. The function $g(q)=h_2(q)+(1-q)\log(M-1)$ has derivative
\begin{equation}
g'(q)=\log\frac{1-q}{q(M-1)}\leq0
\quad\text{for }q\in[1/M,1).
\end{equation}
Thus $q\geq\rho\geq1/M$ implies $H(p)\leq g(\rho)$, proving main-paper Eq.~(28).

\subsection{Energy and concentration measure different properties}
For an illustrative centered vector $s=(-(M-1)a,a,\ldots,a)$ with $a>0$, its energy grows as $M(M-1)a^2$. Under the bounded temperature rule, as $a\to\infty$ its softmax distribution approaches $(0,1/(M-1),\ldots,1/(M-1))$. For $M>2$, the limiting entropy is $\log(M-1)$. This algebraic example explains why high squared energy alone does not specify a nearly one-hot distribution; the exact softmax calculation resolves concentration for a realized score pattern.

For the signed linear coefficients, main-paper Eq.~(29) remains exact for every energy. Its strict monotonicity follows from $b'(z)>0$, independently of Eq.~\eqref{eq:supp_taylor_scope}. This separates the global coefficient-deviation interpretation from the local entropy approximation. The provided scalar energy sample supports the occupancy diagnostic; computing exact realized softmax entropy additionally requires the corresponding score vectors.

\bibliographystyle{IEEEtran}
\bibliography{elinformer}